\documentclass[12pt]{article}

\usepackage{amsmath,amsthm}
\usepackage{amssymb}
\usepackage{amsfonts}
\usepackage{amscd}
\usepackage{color}
\usepackage{subcaption}

\usepackage{tikz}
\usetikzlibrary{arrows.meta, positioning}
\usepackage[all,cmtip,2cell]{xy}
\usepackage{booktabs}
\usepackage{hyperref}
\usepackage{algorithm}
\usepackage{algpseudocode}
\usepackage[utf8]{inputenc}
\usepackage{euler} 
\usepackage{enumitem}
\usepackage{changepage}
\usepackage{soul}

\newtheorem{thm}{Theorem}[section]

\theoremstyle{definition}

\newtheorem{definition}[thm]{Definition}

\newtheorem{example}[thm]{Example}

\newtheorem{observation}[thm]{Observation}

\newcommand{\R}{\mathbb{R}}

\newcommand{\F}{\mathcal{F}}

\newcommand{\cB}{\mathcal{B}}

\newcommand{\cF}{\mathcal{F}}

\newcommand{\beq}{\begin{equation}}
\newcommand{\eeq}{\end{equation}}
\newcommand{\lra}{\longrightarrow}

\begin{document}

\centerline{\Large \bf Sheaf Neural Networks}

\bigskip
\centerline{\Large \bf and biomedical applications}


\bigskip
\centerline{A. Merhab${}^*$, J.-W. Van Looy${}^\star$, P. Demurtas${}^\star$,}

\medskip
\centerline{S. Iotti${}^\star$, E. Malucelli${}^\star$, F. Rossi${}^\star$,
F. Zanchetta${}^\star$, R. Fioresi${}^\star$}

\medskip
\centerline{$*$ University of Ferrara, Italy; $\star$ University of Bologna, Italy}

\bigskip

{\small {\bf Abstract.} The purpose of this paper is to elucidate
the theory and mathematical modelling behind the sheaf neural network (SNN) algorithm and then
show how SNN can effectively answer to biomedical questions
in a concrete case study and outperform the most popular graph
neural networks (GNNs) as graph convolutional networks (GCNs),
graph attention networks (GAT) and GraphSage.
}
\section{Introduction}\label{intro-sec}

Graph Neural Networks (GNNs) \cite{scarselli2008graph}
provide an effective framework for learning
from relational data that may be represented as a graph, 
hence
they are especially fruitful when treating real and challenging datasets
of biomedical significance.
Sheaf Neural Networks (SNNs) are a relatively new family of Deep Learning
(DL) algorithms, generalizing Graph Neural Networks (GNNs), obtained by equipping graphs
with an extra algebraic structure \cite{Bodnar2022, Barbero2022},
namely a {\sl sheaf structure},
which brings along a new enhanced version of the message passing mechanism
and a more fruitful aggregation of node features.

\medskip
The purpose of this work is to show how SNNs can be effectively
used in a concrete case study, involving a biomedical dataset,
 and outperform the most popular graph neural networks
 in a biologically significant classification task.
 This dataset consists of spectroscopic measurements from 
patients affected by osteosarcoma, a highly aggressive primary bone tumor,
characterized by abnormal bone mineralization processes \cite{ritter2010osteosarcoma,
   roschger2008bone}.
Understanding the biochemical alterations in bone tissues,
associated with tumor development is
therefore crucial for improving diagnosis and prognosis. The dataset is
obtained via synchrotron-based X-ray absorption spectroscopy, and details
bone tissue chemical composition at micro and nanoscopic scales, providing 
information on chemical and structural changes associated with disease progression
\cite{rossi2023shedding}.
To better capture the relationships between samples beyond
individual spectral features, we represent the dataset as a graph,
where each node corresponds to a tissue sample and edges are defined by cosine
similarity in the reduced spectral space after principal component analysis (PCA).
Each node is then associated with features, providing information on tissue
properties and it is labelled as tumor/no tumor for the binary
classification task we will perform. 
This graph-based representation provides a natural framework for sheaf neural
networks, with their enhanced representation and classification capabilities.

\medskip
Our work is organized as follows.

\medskip
In Sec. \ref{prelim-sec} we briefly recap the concept of sheaf on a graph
and we compare the definitions appearing in the literature e.g.
cellular sheaves \cite{Curry2013}, with a focus on
the neural network applications \cite{Bodnar2022} introducing also the Laplacian operator
as instrumental to the message passing mechanism.

\medskip
In Sec. \ref{impl-sec} we present the biomedical dataset, and its graph
realization, together with its biological significance.
Then describe our sheaf neural network (SNN) model, together with
the GCN, GAT and GraphSage models and
the training procedure we followed (see also the App. \ref{app} for more
details).

\medskip
In Sec. \ref{res-sec} we present the performance of the SNN model
benchmarked against the most successful GNN models described previously.

\medskip
In Sec. \ref{concl-sec} we present the conclusions and 
we suggest future applications of SNNs to biomedical dataset analysis including clustering
and many graph settings.

\medskip

\section{Sheaves on Graphs} \label{prelim-sec}

In this section we recap the theory of sheaves on graphs, focusing on the
graph neural network applications (see \cite{Fioresi2026}
and refs. therein for the full bibliography). We first give the
usual notion of sheaf on a topological space and then we relate it to
the notion of {\sl cellular sheaf} as it is the one used for the
machine learning applications. Most of this material is well known,
however the treatment of directed/undirected graphs as preordered sets
or topological spaces, deserves some attention, since in the SNN treatments
\cite{Bodnar2022, Barbero2022}
the definitions regarding sheaves on graphs
are not the mainstream ones.

\subsection{Sheaves}

The concept of sheaf, introduced by Leray in \cite{leray} (see also \cite{sheafhistory}), is central in geometry and captures the essence
of the geometrical object it is describing: topological manifolds,
differentiable manifolds or algebraic schemes to cite a few key examples.
Such mathematical objects are defined locally, but exhibit global
behavior which are governed by suitable compatibility conditions. 
The concept of sheaf provides a framework for
describing how local data on these objects can be assembled
to describe global data on them in a coherent way.
For completeness, we briefly recap the key
definitions (see \cite{eh, Tu2007} for a quick introduction to all notions
needed here).

\medskip
\begin{definition}\label{sheaf-def}
Let $X$ be a topological space. A \emph{presheaf} $\mathcal{F}$ on $X$
is an assignment $U \mapsto \mathcal{F}(U)$, where $U$ open in $X$
and $\mathcal{F}(U)$ is a set, and
for each inclusion of open sets $V \subseteq U$, we have a
\emph{restriction map}
    \[
    \rho^U_V : \mathcal{F}(U) \to \mathcal{F}(V),
    \]
satisfying $\rho^U_U = \mathrm{id}$ and $\rho^V_W \circ \rho^U_V = \rho^U_W$
whenever $W \subseteq V \subseteq U$.
We say $\cF$ is a sheaf of groups (rings, vector spaces etc.), if $\cF(U)$ is a
group (ring, vector space etc.).

Let $U \subseteq X$ be an open set and $\{U_i\}_{i \in I}$ an open cover of $U$.
A presheaf $\mathcal{F}$ is called a \emph{sheaf} if it satisfies the
following:

\begin{enumerate}
    \item \textsl{Locality}. If $s,t \in \mathcal{F}(U)$ satisfy
    $s|_{U_i} = t|_{U_i} \quad \text{for all } i \in I$,
    then $s = t$.
    \item \textsl{Gluing}. If for each $i \in I$ there exists a section $s_i \in \mathcal{F}(U_i)$ such that
 $   s_i|_{U_i \cap U_j} = s_j|_{U_i \cap U_j} \quad \text{for all } i,j \in I$,
    then there exists a unique section $s \in \mathcal{F}(U)$ with
    $s|_{U_i} = s_i \quad \text{for all } i \in I$.
\end{enumerate}
\end{definition}

\medskip
\begin{example} {\bf The Sheaf of Differentiable Functions.}
\label{diff-mflds}
Let $M$ be a differentiable manifold \cite{Tu2007}. For each open set $U \subseteq M$, define
\[
\mathcal{C}^\infty(U) = \{ f : U \to \mathbb{R} \mid f \text{ is smooth} \}.
\]
If $V \subseteq U$, the restriction map sends a smooth function
$f \in \mathcal{C}^\infty(U)$ to its restriction $f|_V \in \mathcal{C}^\infty(V)$.
This assignment defines a presheaf of rings on $M$, which is in fact a sheaf.
Smooth functions  defined on an open cover on $M$ can be glued together
to produce a unique smooth function on $M$, \cite{Tu2007}.
\end{example}

\medskip
\begin{observation}\label{b-sheaf}
Given a base $\cB$ for the topology on $X$, a \textit{$\cB$-sheaf} $\cF$ on $X$ is
an assignment $U \mapsto \cF(U)$, for all $U\in \cB$ satisfying the definition of sheaf
when restricted to the open sets in $\cB$, with suitable
modifications (see \cite{eh} Ch. I).
We have that any $\cB$-sheaf on $X$ extends
uniquely to a sheaf on $X$ (Prop. I-12 in \cite{eh}).
\end{observation}

\medskip
We now focus on graphs.
Let $G=(V,E)$ be a (undirected) graph, $V$ the set of its vertices, $E$
its edges. We can define a preorder $\leq$ on $G$ as follows.\footnote{A preorder
differs from an order by not requiring that $x \leq y$, $y \leq x$ implies $x=y$.}
Let $x$, $y$ be vertices or edges of $G$.
We say $x\leq y$ if $x=y$ or $x$ is a vertex of the edge $y$.
As for any finite set with a preorder, we can define a topology by
giving a base for the open sets as follows:
$$
U_p:=\lbrace q\in G\: |\: p\leq q\rbrace \qquad p\in G
$$
where, with an abuse of notation, we see $G$ as the union of its vertices and edges.

\medskip
This topology is called the \textit{Alexandrov topology} \cite{Kosniowski1980, Arenas1999}.
We notice that $U_p$ are \textit{irreducible}, that is there is no smaller
open set containing $p$ and strictly contained in $U_p$.
Irreducible open sets represent an important difference with the continuous setting, where
such open sets do not exist.

\medskip
As one can readily check:
\begin{itemize}
\item $U_v=\lbrace e\in E\: |\: v\leq e\rbrace \, \cup \, \{v\}$,
  that is the open star of $v$, for each vertex $v\in V$,
\item $U_e=\lbrace e\rbrace$, i.e. the edge $e$, without its vertices, for each $e\in E$.
\end{itemize}

\begin{figure}[h]
\begin{center}
\begin{tikzpicture}[scale=.45]

\draw [-] (-6,0) -- (-2,0);
\node (edge) at (-4,0.7) {$e$};

\node (v) at (6,0) {$\bullet$};
\node (vertex) at (6,0.7) {$v$};
\node (a) at (3,-3) {};
\node (b) at (9,-2.5) {};
\node (c) at (3,3) {};
\node (d) at (9,2.5) {};
\node (e) at (9,0) {};
\node (f) at (3,0) {};
\draw [-] (a) -- (v);
\draw [-] (b) -- (v);
\draw [-] (c) -- (v);
\draw [-] (d) -- (v);
\draw [-] (e) -- (v);
\draw [-] (f) -- (v);
\end{tikzpicture}
\end{center}
\caption{Irreducible open sets for the Alexandrov topology on graphs}\label{fig:openstar}
\end{figure}
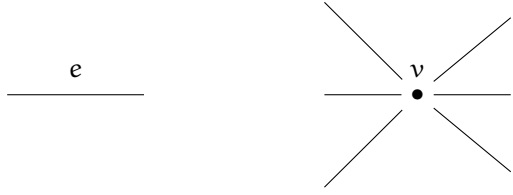

\medskip
\begin{observation} \label{irr-obs}
1. A sheaf $F$ of vector spaces on a graph $G$ viewed as a topological
space as above, is determined by a $\cB$-sheaf $F$ on the irreducible open sets.
Such $F$ consists of:
\begin{itemize}
    \item an assignment of a vector space $F(U_v)$ for each vertex $v\in V_G$.
    \item  an assignment of a vector space $F(U_e)$ for each edge $e\in E_G$.
\item  an assignment of linear maps (restriction maps)
  $F_{v\leq e}:F(U_v)\rightarrow F(U_e)$ for each $v\leq e$ (we
  have $U_e \subset U_v$ whenever $v \leq e$).
\end{itemize}
We leave to the reader the easy check that this assignment gives indeed a $\cB$-sheaf.

\medskip
2. This notion of sheaves on graphs, as described in (1),
is the same as the one introduced by Bodnar
et al. in \cite{Bodnar2022, Curry2013, hg2019} in the context of {\sl cellular sheaves},
which are defined directly on preordered sets associated with regular cell complexes.
Indeed, if we specialize the definition of cellular sheaf to the special
case of a graph, viewed as a regular cellular cell complex, by definition we
have that a cellular sheaf $F$ is an assignment:
$x \lra F(x)$, $F(x)$ a vector space,
for each cell vertex or edge $x$. Moreover if $v \leq e$, i.e. if $v$ is a vertex
of the edge $e$ we have a linear map $F_{v\leq e}:F(U_v)\rightarrow F(U_e)$. Thus
we recover a sheaf, via the remarks in 1., in the usual sense i.e.
as in Def. \ref{sheaf-def}.

\end{observation}

\subsection{Sheaf Laplacian} \label{sec:SGNN}
We define the sheaf Laplacian operator and then we compare
it to the GNN message passing mechanism.

\medskip
For every presheaf of vector spaces $\cF$ on a graph $G=(V,E)$ we define $1$ and $0$ chains
as follows:
$$
C_1(G,\cF):=\oplus_{e\in E} \cF(e), \qquad C_0(G,\cF):=\oplus_{v\in V}\cF(v)
$$
where to ease the notation we use $\cF(x)$ in place of $\cF(U_x)$, for $x$ a vertex or
an edge.

For a given orientation on $G$, we define 
the coboundary operator
\[
\delta : C_0(G,\cF) \lra C_1(G,\cF)
\]
by
\[
    (\delta x)_e := F_{u \leq e}(x_u) - F_{v \leq e}(x_v),
\]
\( x = \{x_v \in \mathcal{F}(v)\}_{v \in V} \in C_0(G,\cF) \), \(\delta x \in C_1(G,\cF) \) and 
$e$ is the edge connecting $u$ and $v$.

We may view $\delta$ as a generalization of the incidence matrix of $G$, that is,
we see $\delta$ as an
operator from the function on vertices to the functions on edges of the graph
(see \cite{godsil} for more details).

\medskip
The \emph{sheaf Laplacian} is defined as
\[
    \mathcal{L}_{\mathcal{F}} = \delta^\top \delta:  C_0(G,F) \lra  C_0(G,F),
\]
This is a positive semi-definite operator acting on $C_0(G,F)$, and its kernel can be
identified with the global sections of the sheaf $\cF$.

When \( \mathcal{F}(v) = \mathbb{R}\), for all \( v \), and all
restriction maps are identities, that is the case when $\F$ equals to the constant sheaf $\underline{\R}$ on $G$, \( \mathcal{L}_{\mathcal{F}} \) reduces to the standard
graph Laplacian \( L = D - A \) \cite{godsil}, where $D$ is the degree matrix and
$A$ the adjacency matrix of $G$  \cite{godsil}.

\subsection{GNN message passing mechanisms}
In order to understand how sheaf Laplacians can improve the design of GNNs,
we need to say more about how GNNs and SNNs are implemented in applications.
To this end, we need to introduce a few notions.
To start with, a \emph{graph $G$ with features $h_v$} consists of the datum of a graph $G$
together with a vector of features $h_v\in\R^n$ for each node $v$ of it. We shall write such datum as $(G,h_G)$ where $\F:V_G\rightarrow \R^n$, $h_G(v):=h_v$.

A typical GNNs and CNNs layer will modify the feature vector
from step $i$ to step $i+1$ as follows:

\begin{equation}\label{eq:gnnlayer}
h^{(i+1)}_v=\sigma_i(\psi_i(GConv_i(G,h^{(i)}_G)_v)+\alpha_i h^{(i)}_vW_i)
\end{equation}
where:
\begin{itemize}
    \item $\sigma_i$ is an activation function (see \cite{DLbookBishop}), that is a multivariate ridge functions.
\item $GConv_i$ 
is a \emph{graph convolution}, i.e. an operator that updates the vector of features according to a so called \emph{message passing} mechanism (see \cite{DLbookBishop}, \cite{FZGDL23}). This is where the key difference between GNNs and SNNs occurs.
    \item $\psi_i$ is usually a regularizing function, $\alpha_i\in\R$ and $W_i$ is a $n_{i}\times n_{i+1}$ matrix of weights (this is known also as a \emph{skip-connection} mechanism, see \cite{DLbookBishop}).
\end{itemize}
\par
A simple example of a graph convolution to use in Eq. \ref{eq:gnnlayer},
is the one where each $GConv_i$ updates the node features using the ordinary
graph Laplacian, that is 
$$
GConv(G,h_G^{(i)})_v=-\sum_{u\in\mathcal{N}(v)}h_u^{(i)}W^{(i)}+ d(v)h_v^{(i)}W^{(i)}
$$
where $\mathcal{N}(v)$ denotes the set of nodes connected with an edge to $v$,
$d(v)=|\mathcal{N}(v)|$ the degree of $v$ and
$W^{(i)}$ is a $n_{i}\times n_{i+1}$ matrix of weights. Fixed an ordering on the nodes of $G$, if we denote as $H^{(i)}\in\R^{|V_G|\times n_i}$ the matrix having as rows
the vectors $h_v^{(i)}$, $v\in V_G$, this convolution can be written in the following compact form, that is often found in literature.
$$GConv_i(G,h_G^{(i)})=AH^{(i)}W^{(i)}$$where $A$ is the adjacency matrix of $G$.
Many of the most popular graph convolutions used in GNNs are a modified version of
this one.
A GNN employing such convolutions can be thought as a discretizing an heat equation diffusing the information stored in each vector of features along the edges at each layer,
see \cite{DLbookBishop, FZGDL23}.

We can see the datum of a graph with features $(G,h_G)$
as the datum of a graph $G$ together with $n$ sections of the constant sheaf $\underline{\R}^n\cong\oplus_{i=1}^n\underline{\R}$. Convolutions as the one just described for the the $i$th layer of a GNN can then be interpreted as operators $C_0(G,\underline{\R})^{n_{i}}\rightarrow C_0(G,\underline{\R})^{n_{i+1}}$ that, using a terminology familiar to the readers working in deep learning, transform sections of the constant sheaf $\underline{\R}$ on $G$ having $n_i$ channels to sections of the constant sheaf $\underline{\R}$ on $G$ having $n_{i+1}$ channels using a mechanism based on the application of $L_{\underline{\R}}$.

The idea of diffusing the information of the node features along the edges of $G$
can be then reinterpreted as the idea of diffusing the information along the stalks
of a constant sheaf (see \cite{Bodnar2022, hg2019}). This idea can then be generalized
to the case when we deal with non-constant sheaves, giving rise to the
so called sheaf neural networks
(see \cite{Bodnar2022} for an explanation and more references).

In SNNs the convolutional layer as in (\ref{eq:gnnlayer}) is then replaced by:

\begin{equation}\label{eq:snnlayer}
h^{(i+1)}_v=\sigma_i(((\Delta_{\F^{(i)}}(I\otimes W_1^{(i)})h_G^{(i)})_vW_2^{(i)}+\alpha_i h^{(i)}_vW_3^{(i)})
\end{equation}
where:
\begin{itemize}
    \item $\sigma_i$ is an activation function.
    \item $\Delta_{\F^{(i)}}$ is the sheaf Laplacian $\mathcal{L}_{\F^{(i)}}$ or the \emph{normalised} sheaf Laplacian $D_{\F^{(i)}}^{-1/2}\mathcal{L}_{\F^{(i)}}D_{\F^{(i)}}^{-1/2}$ (see \cite{Bodnar2022, hg2019}) where, after the identification of $\mathcal{L}_{\F^{(i)}}$ with a $(d\times f_i)\times(d\times f_i)$ matrix, $D_{\F^{(i)}}$ is the block diagonal of $L_{\F^{(i)}}$.
    \item $W_1^{(i)}$, $W_2^{(i)}$and $W_3^{(i)}$ are a $d\times d$, a $f_i\times f_{i+1}$ and a $f_i\times f_{i+1}$ matrices of weights respectively.
    \item $\otimes$ denotes the Kronecker product.
    \item $\alpha_i\in\R$.
\end{itemize}
Explicitly note that in the case $\F^{(i)}=\underline{\R}$ for all $i$, Eq.\ref{eq:snnlayer} coincide with Eq.\ref{eq:gnnlayer} where $\psi_i$ is the identity and $GConv_i$ are the simple Laplacian-based convolutions we described before. SNN are then more flexible than ordinary Laplacian-based GNN as, modeling sheaves that can have a deeper structure than $\underline{\R}$, allow message passing mechanisms that do not treat all the edge relations to be equal allowing the resulting networks to learn complex interactions between nodes that are deeper than the ones contained in an adjacency matrix or in a weighted adjacency matrix (see \cite{Bodnar2022} for a discussion).

\section{Sheaf and Graph Neural Networks}
\label{impl-sec}

In this section we want to illustrate with a concrete case study
the advantage of the sheaf Laplacian message passing mechanism with
respect to the standard ones, as illustrated in the previous section.
We will first briefly introduce the biological dataset and the classification
question we want to solve. Then, we compare the performance of some
sheaf neural network models
and we benchmark them against popular graph neural network
algorithms as GCN, GraphSage, GAT (\cite{GCN, GraphSAGE, GAT18}).

\subsection{The biomedical question}\label{sec:biodatabase}

The dataset we examine is obtained from a series of spectroscopic measurements,
regarding both healthy and tumor bone tissues, of osteosarcoma patients.
Osteosarcoma \cite{ritter2010osteosarcoma, rossi2023shedding}
is a highly aggressive primary bone tumor and it is linked to abnormal mineralization
processes in the bones.
The dataset is obtained via synchrotron-based X-ray absorption near edge structure
(XANES) spectroscopy and describes bone tissue chemical composition at micro-
and nanoscopic scales, providing detailed information on chemical and
structural changes associated with disease progression.
Advanced spectroscopic methods, such as synchrotron-based X-ray spectroscopy
\cite{bertsch2001applications, peyrin2009investigation}
and infrared (IR) spectroscopy
\cite{stuart2000infrared, ng1999infrared},
can be used to provide detailed information on chemical structure.
In particular, synchrotron-based XANES spectroscopy is an X-ray absorption spectroscopy
technique that provides insights into the oxidation state and local coordination
geometry of specific chemical elements constituing the analyzed sample.
In the context of osteosarcoma disease, the chemical environment of calcium is of
paramount interest as this element is a constituent of the mineral fraction of bone tissue.
To obtain the dataset, XANES spectra were collected from 224 points of interest at a
pixel size of 1x1 $\mu$m2 with the incident beam energy being varied between 4.02 keV
and 4.17 keV at 0.3 eV steps across the Ca-K edge
(see \cite{rossi2023shedding} for more details of the experimental
procedure followed to obtain data).
This resulted in a dataset of 224 XANES spectra, each one consisting of $501$ data points.
Each of the 224 samples was labeled as 1 (tumor) or 0 (control), depending on whether the tissue was tumoral or healthy. This produced a labeled dataset with 147 tumor samples and 77 control samples. To this data we assigned the structure of graph with features as follows. First, we perform PCA on the data to reduce the dimension of the spectra to $50$ (in our case, this still accounts for almost 100\% of the variance of the dataset). In the case we consider a train/valid/test split on the samples, PCA is fitted only on the training spectra, and the resulting dimensionality‑reduction model is then applied to all samples. Then, we construct a graph with features $G_{\mathrm{bio}}=(G_{\mathrm{pca}},h_{G_{\mathrm{pca}}})$,
where each node $v$ corresponds to a sample and its
feature vector $(h_{G_{\mathrm{pca}}})_v\in\R^{50}$
is the 50‑dimensional PCA representation of its spectrum. To create the edges, we compute the cosine similarity between every pair of nodes and sort these values from highest to lowest. Edges are then added following this order, starting from the pairs with the highest cosine similarity, until the graph becomes connected. Given this graph, our goal is to train an algorithm that, taking $G_{\mathrm{bio}}$ as input, predicts the label of each node, that is, whether each sample, representing a node, is tumoral or  control.

\subsection{Our GNN and SNN models}\label{sec:models}
We first describe the 3 types of GNN model we tested to benchmark our SNN
and then we go to describe the latter in detail.
Each of GNN models employs layers following Eq. \ref{eq:gnnlayer} and they differ just
for the type of graph convolutions. The
GCN, SAGE and GAT convolution are found in
\cite{GCN, GraphSAGE, GAT18}. We now describe these models in detail.
\par

\medskip
{\bf GCN.} This is a GNN consisting of $q$ layers 
following Eq. \ref{eq:gnnlayer} where:
\begin{itemize}
\item $GConv_i$, for all $i=1,...,q$ is  as in
\cite{GCN}.
    \item if $i\neq q$, $\psi_l$ a normalization function, $W_i$ is equal to the identity matrix if $n_l=n_{l+1}$ and $\alpha_l=1$. $\psi_q$ is the identity function and $\alpha_q=0$.
    \item $\sigma_l$ the ReLU activation function \cite{DLbookBishop} ($i\neq q$) and $\sigma_q$ is the identity function.
\end{itemize}
In addition, dropout \cite{DLbookBishop} is applied after each layer.

\medskip
{\bf GraphSAGE.} This is a GNN consisting of $q$ layers
following Eq. \ref{eq:gnnlayer} where:
\begin{itemize}
    \item $GConv_i$, for all $i=1,...,q$ is the GraphSage convolution \cite{GraphSAGE}, that we implemented using the relative function found in PyG \cite{PyG} (and setting the hyperparameter normalize to be equal to True)
    \item if $i\neq q$, $\psi_l$ a normalization function, $W_i$ is equal to the identity matrix if $n_l=n_{l+1}$ and $\alpha_l=0.05$. $\psi_q$ is the identity function and $\alpha_q=0$.
    \item $\sigma_l$ the ReLU activation function ($i\neq q$) and $\sigma_q$ is the identity function.
\end{itemize}
In addition, dropout is applied after each layer.

\medskip
{\bf GAT.} This is a GNN consisting of $q$ layers 
following Eq. \ref{eq:gnnlayer} where:
\begin{itemize}
    \item $GConv_i$, for all $i=1,...,q$ is the GAT convolution \cite{GAT18}, that we implemented using the relative function found in PyG \cite{PyG} (and setting the hyperparameter normalize to be equal to True)
    \item if $i\neq q$, $\psi_l$ a normalization function, $W_i$ is equal to the identity matrix if $n_l=n_{l+1}$ and $\alpha_l=1$. $\psi_q$ is the identity function and $\alpha_q=0$.
    \item $\sigma_l$ the ELU activation function \cite{DLbookBishop} ($i\neq q$) and $\sigma_q$ is the identity function.
\end{itemize}
In addition, dropout is applied after each layer. We will detail the hyperparameters of the specific models we tested in Appendix \ref{app}.  For all the GNNs we implemented, we have used as regularization function a combination of batch and layer normalization using the functions \texttt{torch\_geometric.nn.BatchNorm} and \texttt{torch.nn.LayerNorm} of PyG \cite{PyG} and Pytorch \cite{Pytorch} respectively.

\medskip
{\bf Sheaf diffusion models.}
The SNN model we tried consist of a SNN having layers following Eq. \ref{eq:snnlayer}. Note that in our experiments the sheaf of each layer is \emph{learnt}, as in \cite{Bodnar2022}, and the sheaves involved at each layer have general, unconstrained restriction maps. In addition, we used the normalized graph Laplacian, we used the same number of channels in each layer and we set $W^{(i)}_3$ as in Eq. \ref{eq:snnlayer} to be equal to the identity matrix if $n_l=n_{l+1}$. We denote this model as SheafGeneral.

\section{Experiments and results}\label{res-sec}
We trained the models described in Section \ref{sec:models} as node (binary)
classifiers, i.e. as models that given as an input the graph with features as
described in Section \ref{sec:biodatabase}, return a label $0$ or $1$ for each node.

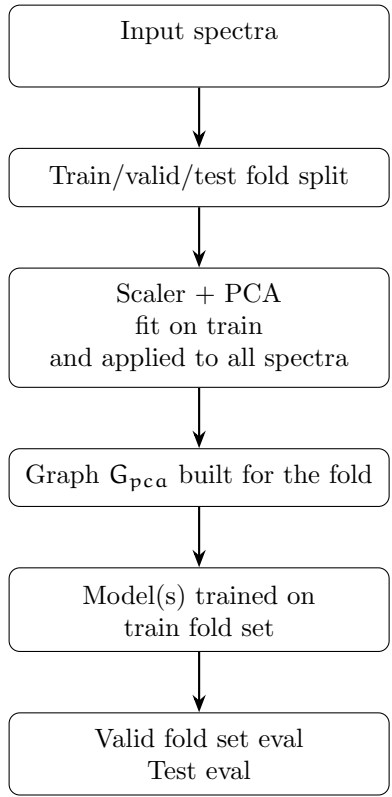
\begin{figure}[h]
\centering
\begin{tikzpicture}[
  node distance=8mm,
  box/.style={
    draw, 
    rounded corners, 
    align=center, 
    inner sep=6pt, 
    minimum width=5cm, 
    fill=white 
  },
  arrow/.style={-{Stealth}, thick}
]

\node[box] (raw) {Input spectra\\};
\node[box, below=of raw] (split) {Train/valid/test fold split};
\node[box, below=of split] (pca) {Scaler + PCA \\ fit on train\\and applied to all spectra};
\node[box, below=of pca] (graph) {Graph $G_{pca}$ built for the fold};
\node[box, below=of graph] (train) {Model(s) trained on\\ train fold set};
\node[box, below=of train] (eval) {Valid fold set eval \\Test eval};

\draw[arrow] (raw) -- (split);
\draw[arrow] (split) -- (pca);
\draw[arrow] (pca) -- (graph);
\draw[arrow] (graph) -- (train);
\draw[arrow] (train) -- (eval);

\end{tikzpicture}
\caption{Training and evaluation workflow for each cross-validation iteration.}\label{fig:workflow}
\end{figure}

We trained our models in a transductive setting (\cite{DLbookBishop}). Recall that under transductive training, for a given train/valid/test split of the nodes, the model receives the full graph topology and all node features, including those of test nodes, but only the labels of the training nodes. Because our dataset contains only 224 samples, we evaluated our algorithms using a 10‑fold stratified cross‑validation (CV \cite{DLbookBishop}), combined with a grid search over the hyperparameters of each of the six models we considered. In a 10‑fold stratified CV, the dataset is partitioned into ten equally sized, non‑overlapping subsets that preserve the original class distribution, resulting in ten distinct train/test splits (the folds). Within each fold, we further split the training portion into an 90/10 train/validation set: PCA is then fit on each reduced training fold set and used to determine the edges $G_{pca}$ for the given fold.
For each hyperparameter configuration in the grid, the corresponding model is trained on the fold’s training set and evaluated on both the validation and test sets.  This entire procedure is repeated $5$ times, each time starting from a different random shuffle of the dataset. For each model type, we then select the best hyperparameter configuration based on the mean validation accuracy across the $5$ repetitions of the ten‑fold cross‑validation. Details on the hyperparameter spaces are provided in Appendix \ref{app}. A summary of the workflow is shown in Fig. \ref{fig:workflow}. Table \ref{table:results} reports two metrics for each model type. The first, Fold Accuracy, is the mean test‑set accuracy across the five repetitions of the ten-fold cross-validation, where in each fold the model was trained on the train fold set using the best hyperparameters during grid search as explained before.
The second metric is the Majority Vote Accuracy. Since each node appears in exactly one test fold and the cross‑validation is repeated 5 times, each model type produces 5 potentially different label predictions for every node. A final label is then assigned to each node by taking the majority vote over these predictions. Using these majority‑vote labels, we compute the Majority Vote Accuracy, reported together with its confidence interval in Table \ref{table:results}. This metric offers a complementary perspective on model performance. It is worth noting that neither our evaluation metrics nor the overall structure of our experimental design are intended to assess the out of sample performance of a final deployed model or to identify the most generalizable solution to the task. Instead, our goal is to compare the architectures under consideration and determine which of them exhibits the greatest expressive power for the problem at hand.

Our models were written using the PyTorch library \cite{Pytorch} in the Python environment. For training our models, we used the Adam optimizer \cite{ADAM} with a learning rate scheduler. The details of the implementation can be found in Appendix \ref{app}.

\begin{table}[t]
\centering
\small
\setlength{\tabcolsep}{4pt}
\renewcommand{\arraystretch}{1.15}
\begin{tabular}{p{3.6cm}c c c}
\toprule
Model & Fold Accuracy (mean$\pm$std) & Majority Vote Acc (95\% CI)  \\
\midrule
SheafGeneral &
0.9779 $\pm$ 0.033 &
0.9866 (0.961--0.995)  \\
\addlinespace

GraphSAGE &
0.9187 $\pm$ 0.061 &
0.9286 (0.887--0.956)  \\
\addlinespace

GAT &
0.7719 $\pm$ 0.080 &
0.7812 (0.723--0.830) \\
\addlinespace

GCN &
0.7603 $\pm$ 0.075 &
0.7634 (0.704--0.814)  \\
\bottomrule
\end{tabular}
\caption{Performances of the models on our dataset.}
\label{table:results}
\end{table}
Our experiments found that SheafGeneral was the model that achieved the best performance, surpassing the performance of all the other models. We also notice that GraphSage was the only vanilla GNN model achieving a performance comparable to the one of the SNNs we tested.

\section{Conclusions} \label{concl-sec}
We have provided a mathematical introduction to
Sheaf Neural Networks showing how they generalize ordinary Laplacian-based graph neural networks. We then showed how SNNs enhanced generalization
capabilities
can effectively answer in a concrete biomedical case study to
relevant biologically grounded questions, outperforming popular
graph neural networks. Future studies will be directed to analyze
datasets consisting of many graphs and enhancing furtherly the SNN
capability towards clustering algorithms and their biomedical applications.

\section{Acknowledgements}
This research was supported by Gnsaga-Indam, by COST Action
CaLISTA CA21109, HORIZON-MSCA-2022-SE CaLIGOLA,
MSCA-DN CaLiForNIA - 101119552, PNRR MNESYS,
PNRR National Center for HPC, Big Data and
Quantum Computing, PNRR SymQuSec and INFN Sezione Bologna.

\section{Contribution}
F. Zanchetta, R. Fioresi contributed in writing the paper and conceptual design
of the research questions. A. Merhab, J.-W. Van Looy, P. Demurtas contributed in writing
both the paper and the code.
S. Iotti, E. Malucelli, F. Rossi contributed with the biological problem
and dataset.

\bibliography{ig-sheafnn}

\appendix
\section{Supplementary material}\label{app}
\subsection{Implementation details}

GCN (and setting the paramter normalize to be equal to True)
We report in Table \ref{tab:final-metrics} the precision, recall, F1‑score, and AUC (Area Under the Curve, see \cite{DLbookBishop}) achieved the best models tested, computed using the same methodology adopted for the Majority Vote Accuracy in Section \ref{res-sec}. The hyperparameter spaces explored in our grid searches are listed in Tables \ref{tab:gcn_grid_current}, \ref{tab:sage_grid_current}, \ref{tab:gat_grid_current}, \ref{tab:sheaf_grid_general_updated}, and the best hyperparameter configurations identified are shown in Table \ref{tab:bestparams}. For each hyperparameter space, all the grid searches performed have been exhaustive. These configurations were used to train the models whose results appear in Tables \ref{tab:final-metrics} and \ref{table:results}. We used the built‑in Adam optimizer provided by PyTorch, together with the \texttt{ReduceLROnPlateau} scheduler from the same library; hyperparameters not included in the grid search were kept at their default values. All trainings were run for 400 epochs, and we did not subdivide the training nodes into batches. For GNN models, early stopping with patience is used. PCA and cosine similarity were computed using the built‑in functions of Scikit‑learn \cite{scikit-learn}. For the graph convolutions employed in our GNNs, we used the implementations available in PyTorch Geometric \cite{PyG}, while the SNN was implemented from scratch in PyTorch. We shall now list all the abbreviations found in the tables below:
\begin{itemize}
    \item hidden\_dim refers to the feature dimension of each layer (the $n_i$ in Eq. \ref{eq:gnnlayer} if $i\neq1,q+1$.
    \item num\_layers is the number of identical layers of the GNN or the SNN considered.
    \item dropout is the dropout rate applied after each layer.
    \item lr is the leraning rate.
    \item weight\_decay is the weight decay parameter of the Adam optimizer.
    \item patience is the patience parameter set to stop training.
    \item sched\_patience is the patience parameter given to the scheduler.
    \item min\_epochs is the minimum number of training epochs before patience can be applied.
    \item grad\_clip is the parameter applied to clip the gradients.
    \item heads is the number of heads. This parameter is specific to GAT layers. In this case Constraint: hidden\_dim must be divisible by heads in the configuration tested.
    \item The hyperparameters f and d of the SheafGeneral are the sheaf dimensions and the number of channels as in Eq- \ref{eq:snnlayer}. 
\end{itemize}

\begin{table}[b]
\centering
\small
\setlength{\tabcolsep}{4pt}
\renewcommand{\arraystretch}{1.15}
\begin{tabular}{p{3.6cm} c c c c c c c c}
\toprule
Model &
Vote Prec. & Vote Rec. & Vote F1 & Vote AUC &
\\
\midrule
SheafGeneral &
1.0000 & 0.9796 & 0.9897 & 0.9969  \\
\addlinespace

GraphSAGE &
0.9712 & 0.9184 & 0.9441 & 0.9483  \\
\addlinespace

GAT &
0.7753 & 0.9388 & 0.8492 & 0.7616 \\
\addlinespace

GCN &
0.7611 & 0.9320 & 0.8379 & 0.6850 \\
\bottomrule
\end{tabular}
\caption{Additional metrics for the final experiment.}
\label{tab:final-metrics}
\end{table}

\begin{table}[b]
\centering
\small
\setlength{\tabcolsep}{6pt}
\begin{tabular}{ll}
\toprule
GraphSAGE hyperparameter & values \\
\midrule
hidden\_dim      & \{16, 32, 64\} \\
num\_layers      & \{2, 3, 4\} \\
dropout          & \{0.1, 0.2, 0.3\} \\
lr               & \{0.001, 0.002, 0.005, 0.01, 0.05\} \\
weight\_decay    & \{1e{-}5, 1e{-}4, 2e{-}4\} \\
\bottomrule
\end{tabular}
\caption{GraphSAGE grid search. Total configurations: $405$.}
\label{tab:sage_grid_current}
\end{table}

\begin{table}[b]
\centering
\small
\setlength{\tabcolsep}{6pt}
\begin{tabular}{ll}
\toprule
GCN hyperparameter & values \\
\midrule
hidden\_dim      & \{32, 64, 128\} \\
num\_layers      & \{2, 3, 4\} \\
dropout          & \{0.1, 0.2, 0.3\} \\
lr               & \{0.002, 0.005, 0.01, 0.05, 0.2, 0.5\} \\
weight\_decay    & \{1e{-}5, 1e{-}4\} \\
patience         & \{80\} \\
min\_epochs       & \{200\} \\
sched\_patience  & \{40\} \\
grad\_clip       & \{1.0, 2.0, 3.0\} \\
\bottomrule
\end{tabular}
\caption{GCN grid search. Total configurations: $972$.}
\label{tab:gcn_grid_current}
\end{table}

\begin{table}[t]
\centering
\small
\setlength{\tabcolsep}{6pt}
\begin{tabular}{ll}
\toprule
GAT hyperparameter & values \\
\midrule
hidden\_dim      & \{32, 64, 128\} \\
num\_layers      & \{2, 3, 4\} \\
dropout          & \{0.1, 0.2, 0.3\} \\
lr               & \{0.001, 0.002, 0.005, 0.01, 0.05, 0.2, 0.5\} \\
weight\_decay    & \{1e{-}5, 1e{-}4\} \\
patience         & \{80\} \\
min\_epochs       & \{200\} \\
sched\_patience  & \{40\} \\
grad\_clip       & \{2.0\} \\
heads            & \{2, 4\} \\
\bottomrule
\end{tabular}
\caption{GAT grid search space. Total configurations: $756$.}
\label{tab:gat_grid_current}
\end{table}

\begin{table}[t]
\centering
\small
\setlength{\tabcolsep}{6pt}
\begin{tabular}{ll}
\toprule
SheafGeneral hyperparameter & values \\
\midrule
d & \{4, 6, 8\} \\
f & \{12, 16, 24\} \\
num\_layers & \{2, 4\} \\
dropout & \{0.1, 0.2, 0.3\} \\
lr & \{0.05, 0.1, 0.2, 0.25\} \\
weight\_decay & \{1e{-}3, 1e{-}4\} \\
activation & ELU \\
patience & 100 \\
grad\_clip & \{1.0\} \\
\bottomrule
\end{tabular}
\caption{Sheaf grid search space focused on general sheaves. Total configurations: $432$.}
\label{tab:sheaf_grid_general_updated}
\end{table}

\begin{table}[t]
\centering
\small
\setlength{\tabcolsep}{4pt}
\renewcommand{\arraystretch}{1.15}
\begin{tabular}{p{3.6cm} p{6.8cm}}
\toprule
Model & hyperparameters \\
\midrule
SheafGeneral &
type=general, $d$=8, $f$=24, $L$=2, act=ELU, drop=0.2, lr=0.01, wd=$10^{-3}$, $\gamma$=0.5, pat=50, clip=0.5 \\
\addlinespace

GraphSAGE &
$h$=32, $L$=2, drop=0.1, lr=0.005, wd=$10^{-4}$, pat=80, min\_ep=100, sched\_pat=40, ls=0.1, clip=2 \\
\addlinespace

GAT &
$h$=32, $L$=4, heads=2, drop=0.1, lr=0.002, wd=$10^{-5}$, pat=80, min\_ep=200, sched\_pat=40, clip=2 \\
\addlinespace

GCN &
$h$=32, $L$=3, drop=0.2, lr=0.002, wd=$10^{-5}$, pat=80, min\_ep=200, sched\_pat=40, clip=1 \\
\bottomrule
\end{tabular}
\caption{Best hyperparameter configurations found with our grid search for each model type.}\label{tab:bestparams}
\end{table}
\subsection{Dataset origin}
Three osteosarcoma patients were chosen for the study, and their bone specimens were
collected during elective surgery at Istituto Ortopedico Rizzoli (IOR) (Bologna, Italy)
prior Ethical Committee approval (744/2019/Sper/IOR) with written informed consent
obtained from each patient.

\end{document}